\documentclass[conference]{IEEEtran}
\IEEEoverridecommandlockouts
\usepackage{cite}
\usepackage{amsmath,amssymb,amsfonts}
\usepackage{algorithmic}
\usepackage{graphicx}
\usepackage{textcomp}
\usepackage{xcolor}
\def\BibTeX{{\rm B\kern-.05em{\sc i\kern-.025em b}\kern-.08em
    T\kern-.1667em\lower.7ex\hbox{E}\kern-.125emX}}

\usepackage{multirow} 
\usepackage{diagbox} 
\usepackage{makecell}
\usepackage{stfloats}
\usepackage{bbding}

\begin{document}

\title{FedDP: Privacy-preserving method based on federated learning for histopathology image segmentation}

\author{
\IEEEauthorblockN{Liangrui Pan\textsuperscript{$\dagger$}}
\IEEEauthorblockA{\textit{College of Computer Science} \\ \textit{and Electronic Engineering} \\
\textit{Hunan University}\\
Chang Sha, China \\
panlr@hnu.edu.cn}
\and
\IEEEauthorblockN{Mao Huang\textsuperscript{$\dagger$}}
\IEEEauthorblockA{\textit{Cancer Research Institute}, \\ \textit{School of Basic Medical Sciences} \\
\textit{Central South University}\\
Chang Sha, China \\
huangmaoo@csu.edu.cn}
\and
\IEEEauthorblockN{Lian Wang}
\IEEEauthorblockA{\textit{College of Computer Science} \\ \textit{and Electronic Engineering} \\
\textit{Hunan University}\\
Chang Sha, China \\
lianwang@hnu.edu.cn}
\and



\IEEEauthorblockN{Pinle Qin}
\IEEEauthorblockA{\textit{School of Computer Science} \\ 
	\textit{North University of China}\\
	Tai yuan, China \\
	qpl@nuc.edu.cn}
\and
\IEEEauthorblockN{Shaoliang Peng*}
\IEEEauthorblockA{\textit{College of Computer Science} \\ \textit{and Electronic Engineering} \\
\textit{Hunan University}\\
Chang Sha, China \\
slpeng@hnu.edu.cn}

\thanks{\textsuperscript{$\dagger$}Equal contribution.}
\thanks{\textsuperscript{*}Corresponding Author.}
}

\maketitle

\begin{abstract}
Hematoxylin and Eosin (H\&E) staining of whole slide images (WSIs) is considered the gold standard for pathologists and medical practitioners for tumor diagnosis, surgical planning, and post-operative assessment. With the rapid advancement of deep learning technologies, the development of numerous models based on convolutional neural networks and transformer-based models has been applied to the precise segmentation of WSIs. However, due to privacy regulations and the need to protect patient confidentiality, centralized storage and processing of image data are impractical. Training a centralized model directly is challenging to implement in medical settings due to these privacy concerns.This paper addresses the dispersed nature and privacy sensitivity of medical image data by employing a federated learning framework, allowing medical institutions to collaboratively learn while protecting patient privacy. Additionally, to address the issue of original data reconstruction through gradient inversion during the federated learning training process, differential privacy introduces noise into the model updates, preventing attackers from inferring the contributions of individual samples, thereby protecting the privacy of the training data.Experimental results show that the proposed method, FedDP, minimally impacts model accuracy while effectively safeguarding the privacy of cancer pathology image data, with only a slight decrease in Dice, Jaccard, and Acc indices by 0.55\%, 0.63\%, and 0.42\%, respectively. This approach facilitates cross-institutional collaboration and knowledge sharing while protecting sensitive data privacy, providing a viable solution for further research and application in the medical field.
\end{abstract}


\section{Introduction}
Histopathological imaging has long been the gold standard for diagnosing cancer~\cite{abels2019computational_1}. H\&E-stained whole slide images (WSIs), which are widely utilized for surgical evaluation, have become an essential point of reference. Accurate segmentation of WSIs enables in-depth analysis of the tumor microenvironment, offering detailed insights~\cite{wang2021hybrid_2}. However, the massive datasets required for pathological image processing involve not only significant computational costs but also raise concerns about patient privacy. In this context, traditional centralized learning faces substantial challenges. Conventional machine learning models typically require training on centralized servers, meaning all pathological image data must be consolidated in one location for processing. However, due to privacy regulations and the need to protect patient confidentiality, the centralized storage and processing of pathological image data have become increasingly impractical~\cite{ogier2023federated_3}.

The core concept of federated learning is to shift the model training process from centralized servers to local devices, allowing each medical device to collaboratively learn while protecting patient privacy. Federated learning enables data from different institutions to remain local, eliminating the need to share original data and only sharing model updates instead. This distributed learning approach offers a novel solution to the privacy and data security challenges faced in the field of pathological image processing~\cite{mcmahan2017communication_4}.

Under the federated learning framework, each medical device can train models locally, learning from the unique features of its local data \cite{pan2024opportunities}. The local model updates are then sent to a central server, where they are aggregated to form a global model. This updated global model is subsequently distributed back to the local devices, continuing iteratively. Throughout this process, the original data remains local, with only the model parameters being shared. This not only helps protect patient privacy but also effectively addresses the diversity and heterogeneity of pathological image data. The collaborative nature of federated learning is also evident during the model update process, as each local device shares model parameters with others, allowing them to complement each other and enhance the overall performance of the model. Moreover, the security of federated learning is ensured. Model updates are transmitted using encryption technologies, safeguarding the security of model parameters during transmission. This is crucial for preventing malicious attacks or theft, especially in the field of pathological image processing, where patient-sensitive information is involved.

Federated learning holds tremendous potential in the field of pathological imaging. It can be applied not only in medical image diagnostics but also extended to medical research. In the diagnostic process of pathological image segmentation, federated learning can enhance model performance, adapt to the data characteristics of different institutions and devices, and better serve patients' diagnostic and treatment needs. In medical research, federated learning facilitates collaboration among multiple institutions, enabling data sharing and mutual benefits~\cite{li2020federated_5,wang2020tackling_6,nguyen2022federated_7,wang2021federated_8,geiping2020inverting_9,hu2022federated_10,lu2022april_11,wen2022fishing_12}.

However, in federated learning, protecting data privacy remains a critical issue~\cite{qi2023differentially_13,wu2022federated_14}. Although federated learning is typically designed to safeguard user data privacy, in some instances, methods such as gradient inversion attacks~\cite{geiping2020inverting_9} can pose risks. These attacks involve analyzing a model's outputs and corresponding gradient information to infer the model's input data, potentially leading to the reconstruction of original data and resulting in privacy breaches~\cite{wei2024multi_15,wu2024concealing_16}. 

Overall, the introduction of federated learning into the field of pathological image processing offers a new perspective on addressing the privacy and data security challenges associated with centralized learning. Its collaborative nature helps enhance the generalizability of models, accommodating the diversity and heterogeneity of pathological image data. Moreover, by decentralizing the handling of data, it preserves the privacy of pathological image data, bringing new hope to the development of the medical field. However, federated learning still faces significant challenges in pathological image processing. This paper addresses the high computational costs and privacy leakage issues by incorporating methods such as differential privacy. Thus, in response to the issues of data silos and privacy breaches in cancer pathological imaging, this paper proposes a privacy-preserving method based on federated learning (FedDP) for segmenting cancer pathology images.

\section{Methods}
\subsection{FedDP overall framework}
\begin{figure*}[h]
	\includegraphics[width=\textwidth]{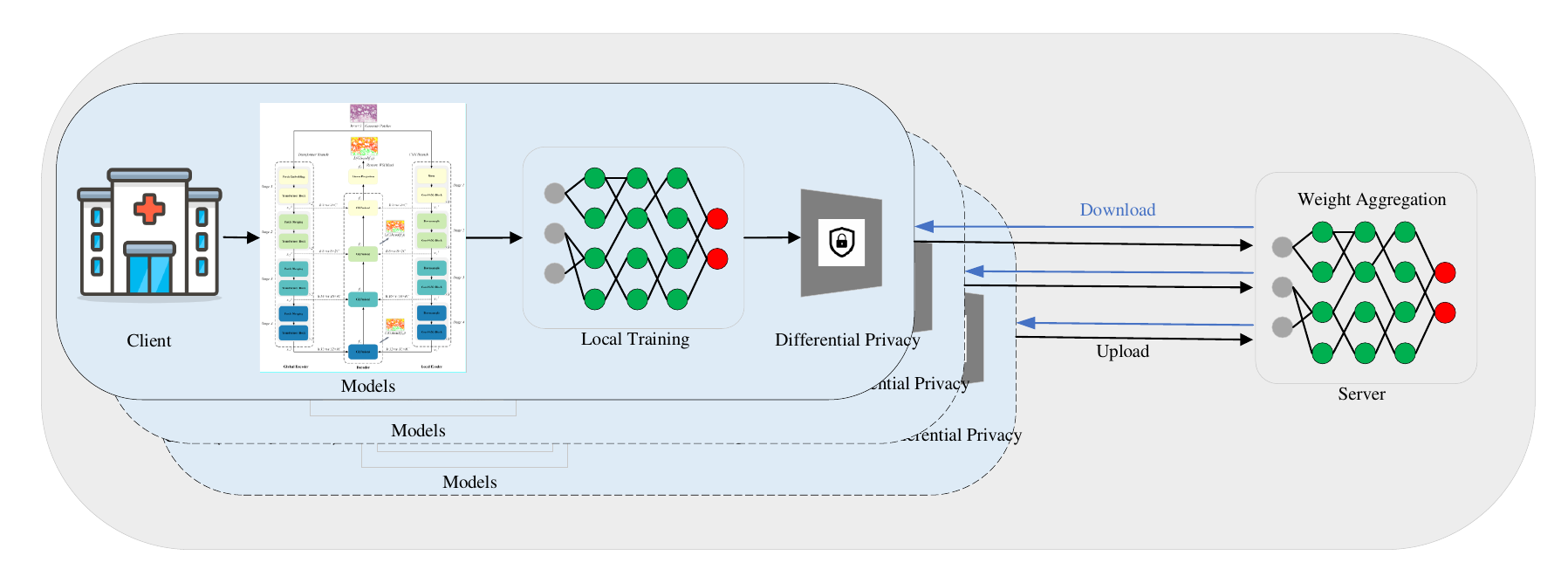}
	\caption{Overall framework of the FedDP model.} 
	\label{FedDPArchitecture}
\end{figure*}
As illustrated in Fig.~\ref{FedDPArchitecture}, the overall framework of the model consists of local training processes on the client side and global aggregation processes on the server side, with the client and server exchanging weight calculations through uploads and downloads \cite{wang2023dhunet,pan2023ldcsf}. For the clients, they must compute differential privacy on the model parameters after local training, and then upload the noise-added parameters to the server. The server, upon receiving the model parameters from all clients, performs a global model aggregation and then distributes the aggregated model back to the clients, ensuring a consistent set of global model parameters.

\subsection{DHUnet}
The architecture of the dual-branch hierarchical global-Local fusion network (DHUnet) is consists of a Swin Transformer~\cite{liu2021swin_20} global encoder, a ConvNeXt~\cite{liu2022convnet_21} local encoder, a decoder, and skip connections. The whole slide image (WSI) is initially divided into smaller patches using a sliding window technique. These patches are processed in parallel by the global and local encoder branches.
In the global encoder, a patch embedding layer segments the patches into non-overlapping sections and maps the feature dimensions to a specified quantity, denoted as C. This reduces the resolution to one-quarter of the original size, resulting in feature dimensions of (w/4, h/4, C). These features undergo consecutive Swin Transformer blocks and patch merging layers to generate hierarchical global feature representations at different scales, such as (w/8, h/8, 2C), (w/16, h/16, 4C), and (w/32, h/32, 8C).
In the local encoder, a stem layer also produces features of size (w/4, h/4, C). These features are further refined through ConvNeXt blocks and downsample layers to generate corresponding hierarchical local feature representations.
Both the global and local encoders in the different stages generate hierarchical feature representations with resolutions similar to conventional convolutional networks like VGG~\cite{simonyan2014very_17} and ResNet~\cite{he2016deep_18}.
In the decoder, the Global-Local Fusion (GLFusion) module and skip connections are utilized to capture coarse-grained global information and fine-grained local information from corresponding levels of the dual branches during the upsampling process. Additionally, a Cross-scale Expand Layer module is introduced to enable upsampling with the same center but different scales, enhancing segmentation results.
Finally, a Linear Projection layer is applied to the upsampled features to output pixel-level segmentation for the small patches, and the segmentation results of all sliced blocks are merged to complete the segmentation of the entire WSI.

In the Swin Transformer global encoder, the patch embedding layer (PE) initially converts the input RGB image $x\in {{\text{R}}^{h\times w\times 3}}$ into non-overlapping patches of size 4x4, similar to the ViT~\cite{dosovitskiy2020image_19}. Each patch is treated as a token, and its features are concatenated. Subsequently, the features with dimensions 4x4x3 are projected onto an arbitrary dimension C. To facilitate the exchange of information between neighboring windows, a series of Swin Transformer blocks (STBs) is employed. These STBs enable effective communication and interaction among the tokens in neighboring windows while preserving the total number of tokens. In this section, the number of Swin Transformer blocks per stage is adjusted from (2, 2, 6, 2) to (2, 2, 2, 2). To differentiate features in the local encoder, the subscript '$g$' is used. Thus, the process of computing global features in Stage 1 can be described as follows:
\begin{equation}
	\centering
	x_{1}^{g}=ST{{B}_{2}}(PE(x)),\ \ \ \ x_{1}^{g}\in {{\mathbb{R}}^{\frac{h}{4}\times \frac{w}{4}\times C}}
\end{equation}

To generate hierarchical feature representations, the Patch Merging (PM) layer plays a crucial role in the network's deepening process. It concatenates the features from every group of 2x2 adjacent patches and utilizes a mapping layer to downsample the resolution by a factor of two while doubling the channel dimensions. This process contributes to the creation of hierarchical feature representations. Consequently, the process of computing global features for Stages 2, 3, and 4 can be described as follows:
\begin{equation}
	\centering
	\begin{split}
	& x_{i}^{g}=ST{{B}_{2}}(PM(x_{i-1}^{g})) \\ 
	& x_{i}^{g}\in {{\mathbb{R}}^{\frac{h}{{{2}^{i+1}}}\times \frac{w}{{{2}^{i+1}}}\times {{2}^{i-1}}C}},\ \ \ \ i=2,3,4 \\ 
	\end{split}
\end{equation}

In the ConvNeXt local encoder, similar to the global encoder process, the input image $x\in {{\text{R}}^{h\times w\times 3}}$ first goes through the Stem layer. Following that, $x_{\text{stem}}^{\text{l}}$ undergoes processing in multiple ConvNeXt (CNBs) blocks to facilitate local feature fusion. In this particular section, the number of ConvNeXt blocks per stage has been adjusted from (3, 3, 9, 3) to (3, 3, 3, 3). To differentiate features in the local encoder, the subscript '$l$' is used. The computation process for local features in Stage 1 can be described as follows:
\begin{equation}
	\centering
	\begin{split}
		x_{1}^{l}=CN{{B}_{3}}(Stem(x)),\ \ x_{1}^{l}\in {{\mathbb{R}}^{\frac{h}{4}\times \frac{w}{4}\times C}}
	\end{split}
\end{equation}

Similar to the patch embedding layer, the downsampling layer (DS) utilized between each stage employs a convolution operation with a kernel size and stride of 2. This operation achieves a twofold reduction in resolution and doubles the channel dimensions. Additionally, a Layer Normalization (LN) layer is applied. Hence, the computation of local features for Stages 2, 3, and 4 can be summarized as follows:
\begin{equation}
	\centering
	\begin{split}
		& x_{i}^{l}=CN{{B}_{3}}(DS(x_{i-1}^{l})), \\
		& x_{i}^{l}\in \ {{\mathbb{R}}^{\frac{h}{{{2}^{i+1}}}\times \frac{w}{{{2}^{i+1}}}\times {{2}^{i-1}}C}},\ \ \ \ i=2,3,4 \\
	\end{split}
\end{equation}

In the GLFusion decoder, to ensure the efficient integration of hierarchical features from the dual encoder branches, we propose a novel Global-Local Fusion (GLFusion) decoder. The GLFusion decoder aims to restore spatial resolution and generate segmentation results. For faster operational speed and enhanced performance, the GLFusion module initially uses additive operations to merge global and local features from the same stage. The result is then connected to the previous module via skip connections (SC), using convolution, batch normalization, and ReLU activation functions to align the output size with the input channel dimensions. Subsequently, an upsampling process, utilizing a Cross-scale Expand Layer which operates inversely to the downsampling and patch merging layers, is implemented. This upsamples the resolution by a factor of two while halving the channel dimensions or by a factor of four while maintaining the channel dimensions. It is accomplished by employing multiple transposed convolutions with identical strides but varying kernel sizes. The features generated by these convolutions are then concatenated along the channel dimension. This approach enables patches with the same center but different scales to achieve diverse receptive fields. The computation process for $GLFusion_4$ can be described as follows:
\begin{equation}
	\centering
	\begin{split}
		& {{f}_{4}}=concat[transpos{{e}_{t}}(SC(x_{i}^{l}+x_{i}^{g}))_{t=1}^{M}],\quad  \\ 
		& {{f}_{4}}\in {{\mathbb{R}}^{\frac{h}{8}\times \frac{w}{8}\times 4C}} \\ 
	\end{split}
\end{equation}
Where $f_4$ represents the output of the $GLFusion_4$ module. The computation processes for $GLFusion_2$ and $GLFusion_3$ can be described as follows: 
\begin{equation}
	\centering
	\begin{split}
		& {{f}_{i}}=concat[transpos{{e}_{t}}(SC(x_{i}^{l}+x_{i}^{g},{{f}_{i+1}}))_{t=1}^{M}],\quad  \\ 
		& {{f}_{i}}\in {{\mathbb{R}}^{\frac{h}{{{2}^{i}}}\times \frac{w}{{{2}^{i}}}\times {{2}^{i-2}}C}},\ \ \ \ i=2,3 \\ 
	\end{split}
\end{equation}

The computation process for GLFusion1 can be described as follows:
\begin{equation}
	\centering
	\begin{split}
		& {{f}_{1}}=concat[transpos{{e}_{t}}(SC(x_{1}^{l}+x_{1}^{g},{{f}_{2}}))_{t=1}^{M}],\quad  \\ 
		& {{f}_{1}}\in {{\mathbb{R}}^{h\times w\times C}} \\ 
	\end{split}
\end{equation}

Finally, a linear projection layer (LP) is applied to the output $f_1$ of the $GLFusion_1$ module to obtain pixel-level segmentation results for the sliced blocks: 
\begin{equation}
	\centering
	Segmentation(x)=LP(f_1)
\end{equation}
The segmentation results of all sliced blocks are merged to reconstruct the final distribution of the tumor in the whole slide image (WSI).

\subsection{Client-side training}
This chapter, based on the FedAvg strategy, addresses gradient privacy leakage by incorporating differential privacy techniques. It assumes that there are $K$ clients participating in the training, with each client $k$ possessing their own local dataset ${{D}_{k}}$ and sharing the same initial global model parameters ${{\theta }_{0}}$.
\begin{itemize}
	\item [1.] Global model Initialization ${\theta}_{0}\sim Initialization$, $Initialization$ is the initialization method of the global model parameters.
	\item [2.] Client Selection: A portion $K$ is randomly selected from $C$ clients as participants in this round.
	\item [3.] Model Distribution: The central server distributes the global model parameters $\theta $ to the selected participants.
	\item [4.] Local training: Each client $c\in C$ uses the local dataset ${{D}_{c}}$ to perform local model training and update the gradient. After the $t$-th round of distributed training, client c performs $E$ local updates locally to obtain the local model parameters $\theta _{t+1}^{c}$:
	\begin{equation}
		\centering
		\theta _{t+1}^{c}=\text{ClientUpdate}(\theta _{t}^{c},{{D}_{c}},E),\ \ t\ge 0
	\end{equation}
	Among them, $\text{Client Update}$ is the function of updating model parameters locally on client $c$, and the gradient descent method is used to implement the training of deep learning model: 
	\begin{equation}
		\centering
		{{\theta }_{t+1}}={{\theta }_{t}}+\frac{1}{k}\sum\limits_{i=0}^{k}{\Delta {{\theta }^{i}}}
	\end{equation}
	
\end{itemize}

\subsection{Differential privacy}
Differential Privacy (DP) is a privacy-preserving technique designed to provide meaningful data analysis results while ensuring the protection of individual privacy during statistical analysis or data mining. Differential privacy achieves this by introducing noise into the computation process, making it impossible to infer specific individual information from the output.

Noise Addition: To protect privacy, clients need to introduce noise when computing gradients. This noise can be random and must satisfy the conditions of differential privacy. Typically, Laplace noise or Gaussian noise is used for differential privacy computation.

Laplace Noise: Participants generate Laplace noise based on the differential privacy parameter $\varepsilon $ and sensitivity $\Delta f$. The probability density function of the Laplace distribution is given by $\rho (x)=\frac{1}{2\varepsilon }{{e}^{-\frac{\left| x \right|}{\varepsilon }}}$, where $x$ represents the noise value. The scale of the noise is determined by the sensitivity of the gradient, i.e., the noise scale is $\frac{\Delta f}{\varepsilon }$.

Gaussian Noise: Participants can generate noise using a Gaussian distribution, where the probability density function of the Gaussian noise is $\rho (x)=\frac{1}{\sqrt{2\pi }\sigma }{{e}^{-\frac{{{x}^{2}}}{2{{\sigma }^{2}}}}}$, with $x$ being the noise value and $\sigma $ being the standard deviation.

In the privacy computing process, the model training update phase uses the update function after adding Gaussian noise:
\begin{equation}
	\centering
	{{\theta }_{t+1}}={{\theta }_{t}}+\frac{1}{k}(\sum\limits_{i=0}^{k}{\Delta {{\theta }^{i}}}/\max (1,\frac{{{\left| \Delta {{\theta }^{i}} \right|}_{2}}}{C})+N(0,{{\sigma }^{2}}{{C}^{2}}I))
\end{equation}
Where $\sigma $ and $C$ are hyperparameters in the privacy computing strategy. The conditions for the Gaussian mechanism to satisfy $(\varepsilon ,\delta )$-differential privacy are:
\begin{equation}
	\centering
	\varepsilon =\frac{\vartriangle f}{\sigma }
\end{equation}
where $\vartriangle f$ represents the sensitivity, indicating the maximum change in the function's output on neighboring datasets. For gradient computation, sensitivity can be controlled by clipping the gradients. Assuming the clipping threshold is $C$, we have:
\begin{equation}
	\centering
	\vartriangle f\le C
\end{equation}
Thus, the relationship between the privacy budget $\varepsilon $ and the noise standard deviation $\sigma $ is:
\begin{equation}
	\centering
	\varepsilon =\frac{C}{\sigma }
\end{equation}

It is important to note that when adding Gaussian noise, attention should be paid to the scale of the noise, specifically the choice of the standard deviation. A larger standard deviation increases the magnitude of the noise, providing stronger privacy protection, but it may also impact the accuracy and usability of the data. Therefore, in practical applications, a suitable trade-off and adjustment are necessary to balance the requirements of privacy protection and data utility.

\subsection{Server-side aggregation}
After completing local training, the client sends its updated or encrypted model parameters $\theta _{t+1}^{(c)}$ to the server. Once the server receives updates from all clients, it performs global parameter aggregation using a weighted average method:
\begin{equation}
	\centering
	{{\theta }_{t+1}}=\frac{1}{C}\sum\limits_{c=1}^{C}{\theta _{t+1}^{(c)}}
\end{equation}

Once the server completes the aggregation, it distributes the global parameters to each client. The process of client training and applying differential privacy is iteratively repeated until a stopping condition is satisfied. This stopping condition can be defined as reaching the maximum number of iterations or attaining model convergence.

\section{Experimental}
\subsection{Experiment details}
The experiments in this study are conducted using Python 3.7 and PyTorch 1.7.0. Data augmentation techniques such as horizontal flipping, vertical flipping, and random rotations between -90° and 90° are employed to enhance the diversity of data positions for all training models. Additionally, to enhance the network's resilience to variations in color, random hue-saturation-value (HSV) transformations are applied. The dataset is then divided randomly into training and testing sets, with a ratio of 0.7:0.3. The models are trained using 5-fold cross-validation on an NVIDIA Tesla V100 GPU (16GB VRAM). The optimal parameters for the models are determined by averaging the performance on the validation sets. The input size for the model is set to 224×224, and the maximum number of training epochs for the WSSS4LUAD dataset is set to 150 to ensure loss convergence.During the training phase, a standard SGD optimizer is used for backpropagation network optimization, with a default batch size of 24. The experiment included 5 local clients, with each client performing 5 local iterations, and the server performing 150 rounds of global integration. The CPU used is a 16-core Intel(R) Xeon(R) Bronze 3106 CPU @ 1.70GHz. Additionally, differential privacy parameters and are set to 0.5 and 0.05, respectively.
\subsection{Datasets}
Experiments on the Lung Cancer WSSS4LUAD Dataset includes labels for tumors, stroma, and normal tissues, and is publicly provided by the WSSS4LUAD 2021 Challenge\footnote{https://wsss4luad.grand-challenge.org/WSSS4LUAD/}. The dataset comprises 23 large patches selected from Guangdong Provincial People's Hospital and The Cancer Genome Atlas (TCGA). The patches have resolutions ranging approximately from 1500×5000 to 1500×5000. Additionally, 20 whole slide images (WSIs) are collected from TCGA (one WSI per patient), combining to form the lung cancer WSSS4LUAD dataset. The dataset undergoes processing using a sliding window approach with a resolution of 1000×1000 and a stride of 500. Subsequently, the images are resized to 224×224 to match the model's scale.

\section{Results}
\subsection{FedDP achieve state-of-the-art performance}
As shown in Table~\ref{tab_feddp}, under the baseline framework based on federated learning, DHUnet demonstrates significant advantages compared to other CNN-based and Transformer-based models. On the WSSS4LUAD lung cancer dataset, its Dice, Jaccard, and Acc scores reach 85.23\%, 76.26\%, and 90.71\%, respectively, showcasing outstanding performance. These metrics are widely recognized in image segmentation tasks and are crucial for evaluating model performance. Although the performance of all models decreased compared to methods without federated learning, the security of data and models cannot be ensured without it.

\begin{table}[ht]
	 \renewcommand{\arraystretch}{1.5} 
	\centering
	\caption{Segmentation performance of FedDP method on lung cancer WSSS4LUAD dataset.} 
	\label{tab_feddp}
	\resizebox{0.5\textwidth}{16mm}{
		\begin{tabular}{ccccccccccc}
			\Xhline{1.5pt}
			\multirow{2}{*}{Models} & \multicolumn{3}{c}{None} & \multicolumn{3}{c}{Federate Learning} & \multicolumn{3}{c}{\textbf{FedDP}} \\
			~ & Dice & Jaccard & Acc & Dice & Jaccard & Acc & Dice & Jaccard & Acc \\ \hline
			FCN & 92.31 & 85.72 & 97.24 & 78.40 & 68.40 & 87.08 & 72.15 & 62.56 & 80.03  \\ 
			Unet & 92.71 & 86.43 & 97.40 & 79.09 & 69.05 & 86.82 & 78.45 & 68.28 & 86.86  \\ 
			DeeplabV3 & 92.71 & 86.44 & 97.47 & 83.84 & 74.50 & 91.24 & 51.52 & 47.02 & 49.39  \\ 
			ConvNeXt & 92.89 & 86.75 & 97.47 & 83.08 & 73.48 & 89.37 & 82.54 & 72.99 & 89.14  \\ 
			SwinUnet & 91.76 & 84.84 & 97.25 & 84.04 & 74.59 & 90.04 & 83.98 & 74.47 & 89.96  \\ 
			TransFuse & 90.47 & 82.72 & 96.76 & 85.29 & 76.21 & 90.97 & 84.64 & 75.30 & 90.71  \\ 
			DHUnet & \textbf{93.07} & \textbf{87.04} & \textbf{97.52} & \textbf{85.78} & \textbf{76.89} & \textbf{91.29} & \textbf{85.23} & \textbf{76.26} & \textbf{91.09}  \\ 
			\Xhline{1.5pt}
		\end{tabular}
	}
\end{table}

Notably, DHUnet manages to maintain high accuracy even with the incorporation of differential privacy through FedDP, resulting in minimal Dice, Jaccard, and Acc losses of only 0.55\%, 0.63\%, and 0.42\%, respectively, while still maintaining a leading edge in segmentation performance. Compared to the best-performing Transformer-based model, TransFuse, DHUnet improves Dice, Jaccard, and Acc scores by 0.59\%, 0.96\%, and 0.38\%, respectively. When compared to the best-performing CNN-based model, ConvNeXt, it shows improvements of 2.69\%, 3.27\%, and 1.95\% in Dice, Jaccard, and Acc metrics, respectively. These enhancements are likely attributed to the design of the global-local fusion decoder and skip connections in the DHUnet model.

\begin{figure}[ht]
	\includegraphics[width=0.5\textwidth]{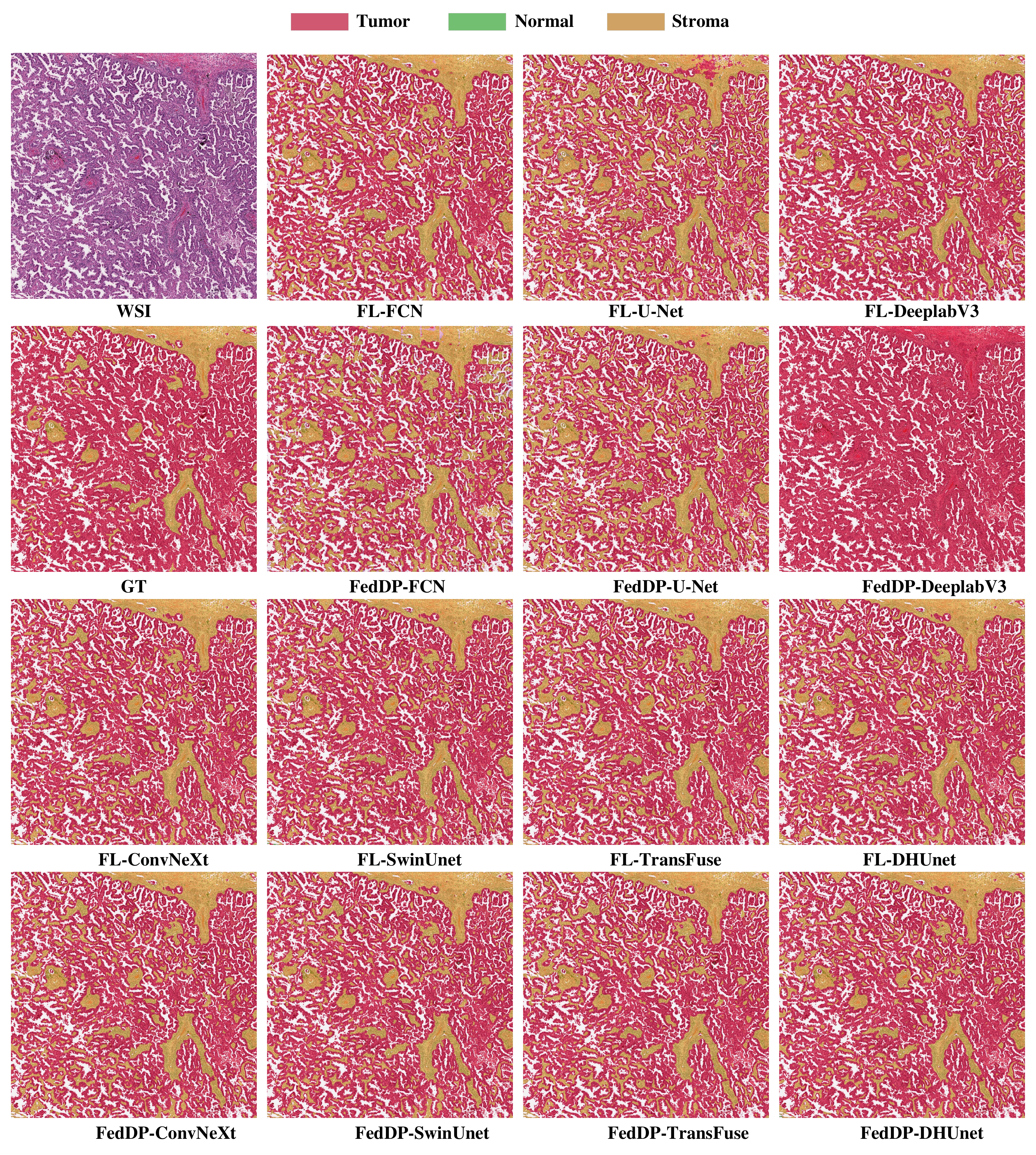}
	\caption{Visualization results of Federated Learning (FL) and FedDP methods.} 
	\label{Result1}
\end{figure}

In the visualization results shown in Fig.~\ref{Result1}, DHUnet also demonstrates excellent performance. Under the federated learning framework, DHUnet continues to exhibit high segmentation quality in image segmentation tasks, maintaining a high degree of similarity, overlap, and accuracy with the ground truth. This highlights its robustness and generalization capabilities. Among all models, the FedDP method showcases the best segmentation performance. Therefore, FedDP not only accurately segments pathological images but also ensures data privacy and model security.

\begin{figure}[bp]
	\includegraphics[width=0.5\textwidth]{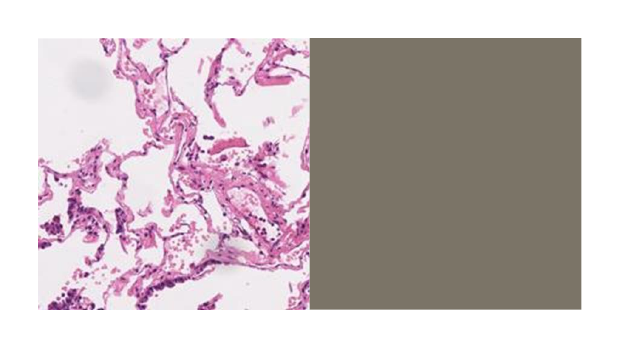} 
	\caption{The left side shows the lung cancer pathology image, and the right side shows the image reconstructed after inversion attack.} 
	\label{Result2}
\end{figure}

As illustrated in Fig.~\ref{Result2}, inversion attacks are unable to reconstruct the original pathological images. Compared to traditional federated learning methods, the introduction of differential privacy technology significantly enhances the privacy of data distribution and effectively prevents data leakage and privacy breaches. This provides users with more reliable data protection measures. Additionally, the application of this technology increases users' trust in data-sharing schemes, as they can share data with greater confidence, without fear of privacy being compromised or misused. Therefore, the FedDP method has broad application prospects in the fields of data sharing and privacy protection, offering vital support for building secure and reliable data-sharing solutions.

In summary, FedDP demonstrates outstanding performance within the federated learning framework, maintaining high levels of accuracy even with the inclusion of differential privacy. FedDP achieves notable results in performance metrics such as Dice, Jaccard, and Acc, while also enhancing the privacy of data distribution through differential privacy technology. This strengthens the security and privacy protection of user data. The analysis of visualization results further confirms FedDP's superiority in image segmentation tasks, showcasing its robustness and generalization capabilities. Consequently, FedDP provides strong support and assurance for data sharing and privacy protection.

\subsection{Ablation Experiment}
\begin{table}[ht]
	\renewcommand{\arraystretch}{1.5} 
	\centering
	\caption{Effect of noise scale on the model.} 
	\label{tab_ablation}
	\resizebox{0.5\textwidth}{16mm}{
		\begin{tabular}{ccccccccccc}
		\Xhline{1.5pt}
		\multirow{2}{*}{\diagbox{$\sigma$}{$C$}} & \multicolumn{3}{c}{0.1} & \multicolumn{3}{c}{0.3} & \multicolumn{3}{c}{0.5} \\ 
		~ & Dice & Jaccard & Acc & Dice & Jaccard & Acc & Dice & Jaccard & Acc \\ \hline
		0.05 & 83.36 & 74.07 & 89.46 & 84.79 & 75.86 & 90.21 & 85.23 & 76.26 & 91.09  \\ 
		0.15 & 83.55 & 74.37 & 89.46 & 84.64 & 75.64 & 90.14 & 85.02 & 76.03 & 90.48  \\ 
		0.25 & 80.51 & 70.53 & 88.39 & 84.59 & 75.71 & 90.00 & 83.21 & 64.28 & 88.73  \\ 
		0.35 & 79.87 & 69.64 & 88.29 & 84.49 & 75.61 & 89.87 & 83.20 & 64.12 & 88.70  \\ 
		\Xhline{1.5pt}
		
	\end{tabular}	
}

\end{table}
As shown in Table~\ref{tab_ablation}, ablation experiments reveal that a larger standard deviation increases the magnitude of noise, providing stronger privacy protection but potentially impacting data accuracy. Conversely, increasing the C value enhances the overall accuracy of the model. Therefore, this study selects the optimal values as input parameters for the model to ensure the best possible performance.

\section{Conclusion}
The challenge of training deep neural network models with good generalization performance is significantly heightened due to the difficulty of sharing image data across different medical institutions. To address this issue, this chapter proposes a privacy-preserving method based on federated learning. Additionally, the chapter addresses the risk of sensitive data leakage through the transmission of model updates during the training process. Federated learning allows each medical institution to train models locally without the need to share sensitive image data. Furthermore, we employ differential privacy techniques to enhance data privacy protection. Differential privacy introduces noise into model updates, making it impossible for attackers to infer the contributions of individual samples, thereby safeguarding the privacy of the training data. Experimental results demonstrate that our proposed method has minimal impact on model accuracy while effectively protecting data privacy. This means that it is possible to ensure the privacy and security of medical image data without significantly sacrificing accuracy. Therefore, the FedDP method has broad application prospects in the fields of data sharing and privacy protection, providing vital support for building secure and reliable data-sharing solutions.

\section*{Acknowledgment}
This work was supported by Hunan Province Graduate Research Innovation Project CX20240450; National Key R\&D Program of China 2023YFC3503400, 2022YFC3400400; NSFC-FDCT Grants 62361166662; The Innovative Research Group Project of Hunan Province 2024JJ1002; Key R\&D Program of Hunan Province 2023GK2004, 2023SK2059, 2023SK2060; Top 10 Technical Key Project in Hunan Province 2023GK1010; Key Technologies R\&D Program of Guangdong Province (2023B1111030004 to FFH). The Funds of State Key Laboratory of Chemo/Biosensing and Chemometrics, the National Supercomputing Center in Changsha (http://nscc.hnu.edu.cn/), and Peng Cheng Lab.
\bibliographystyle{IEEEtran}
\bibliography{REFERENCES}

\end{document}